%% file: main.tex
\documentclass{article}

\usepackage[final]{neurips_2024}

\usepackage[utf8]{inputenc} % allow utf-8 input
\usepackage[T1]{fontenc}    % use 8-bit T1 fonts
\usepackage{hyperref}       % hyperlinks
\usepackage{url}            % simple URL typesetting
\usepackage{booktabs}       % professional-quality tables
\usepackage{amsfonts}       % blackboard math symbols
\usepackage{nicefrac}       % compact symbols for 1/2, etc.
\usepackage{microtype}      % microtypography
\usepackage{xcolor}         % colors
\usepackage{graphicx} % self
\usepackage{array} % self
\usepackage{multirow} % self
\usepackage{makecell} % self
\usepackage{caption} % self

\input{section/macros}

\title{TimeFound: A Foundation Model for Time Series Forecasting}

\author{
Congxi Xiao$^{1,2,\dagger}$, Jingbo Zhou$^{1,*}$, 
 Yixiong Xiao$^1$, Xinjiang Lu$^1$, Le Zhang$^1$, Hui Xiong$^{3,}$\thanks{Jingbo Zhou and Hui Xiong are corresponding authors. $^\dagger$This work was done when the first author was an intern at Baidu Research under the supervision of Jingbo Zhou}
 \\
$^1$ Business Intelligence Lab, Baidu Research\\
$^2$ University of Science and
Technology of China\\
$^3$ The Hong Kong University of Science and
Technology (Guangzhou) \\
  \texttt{xiaocongxi@mail.ustc.edu.cn,} \\
  \texttt{\{zhoujingbo, xiaoyixiong, luxinjiang, zhangle09\}@baidu.com,} \\
  \texttt{xionghui@ust.hk} \\
}

\begin{document}

\maketitle

\begin{abstract}
We present \mymodel, an encoder-decoder transformer-based time series foundation model for out-of-the-box zero-shot forecasting. 
To handle time series data from various domains,
\mymodel employs a multi-resolution patching strategy to capture complex temporal patterns at multiple scales.
We pre-train our model with two sizes (200M and 710M parameters) on a large time-series corpus comprising both real-world and synthetic datasets.
Over a collection of unseen datasets across diverse domains and forecasting horizons, our empirical evaluations suggest that \mymodel can achieve superior or competitive zero-shot forecasting performance, compared to state-of-the-art time series foundation models.

\end{abstract}

\section{Introduction}
\input{section/intro}

\section{Related Work}
\input{section/related}

\section{Problem Definition}
\input{section/preliminary}

\section{Method}
\input{section/method}

\section{Experiments}
\input{section/experiment}

\section{Conclusion}
\input{section/conclusion}

\bibliographystyle{plainnat}
\bibliography{reference}

%%%%%%%%%%%%%%%%%%%%%%%%%%%%%%%%%%%%%%%%%%%%%%%%%%%%%%%%%%%%

% \appendix

% \section{Appendix / supplemental material}

\end{document}

%% file: section/macros.tex
\usepackage{xspace}
\usepackage{amsmath}
\usepackage{bm}

\newcommand{\mymodel}{{TimeFound}\xspace}

%% file: section/intro.tex
Time series forecasting \citep{hyndman2018forecasting} plays a crucial role in industrial applications and scientific research in various domain \citep{chen2012bayesian, zhou2015smiler, deb2017review, karmy2019hierarchical, kaushik2020ai, zhu2023energy, ji2023spatio,zhou2024sdwpf}, such as energy, retail, finance, manufacturing, and healthcare.
In recent years, data-driven deep learning models have demonstrated remarkable success in time series forecasting \citep{salinas2020deepar, sen2019think, zhou2021informer, zeng2023transformers, nietime, chen2021autoformer}, surpassing traditional statistical models like ARIMA. 
Despite their impressive effectiveness, a major limitation of deep forecasters is the heavy reliance on substantial task-specific training data. 
This restricts their ability to generalize to diverse forecasting scenarios, especially on those where data is scarce and insufficient to support additional training, i.e., necessitating zero-shot forecasting.

Witnessing the recent advance of language foundation models (i.e. Large Language Models, LLMs), researchers have been inspired to develop time series foundation models that are generalizable to a broad range of forecasting scenarios \citep{liang2024foundation}.
Following the paradigm of building LLMs, recent studies (e.g., \citep{dasdecoder, ansari2024chronos, liutimer, shi2024time, garza2023timegpt, woo2024unified}) collect a large scale of heterogeneous time series data from multiple domains to pre-train the time series foundation model in a self-supervised manner. 
After learning to capture common temporal patterns from extensive and diverse range of time series data, these models achieve superior forecasting performance and show promising generalization capabilities on unseen data without any training.
Though still in its early stages, the development of time series foundation models marks a paradigm shift in forecasting, moving toward a more adaptable solution for building a universal forecaster across diverse data distributions.

In this work, we continue exploring the development of effective foundation models for time series forecasting, and propose \mymodel, a transformer-based time series foundation model.
In terms of the architecture, we employ an encoder-decoder design for time series modeling and forecasting, where the encoder enables contextual understanding of historical trends while the decoder maintains the causal future prediction.
To tokenize time series data, we adopt a multi-resolution patching method that performs multiple divisions with different patch sizes, rather than fix-size patching. 
This design is driven by the need for a foundation model to deal with the time series across various domains with distinct dynamics and frequencies. In addition, even the same time series can exhibit diverse variations and fluctuations at different temporal scales \citep{ding2024drformer, chenpathformer}. Our approach facilitates the capture of temporal patterns at multiple scales, enhancing the model’s ability to handle diverse forecasting scenarios.

We pre-train \mymodel in two size (\textit{\mymodel-Base}-200M and \textit{\mymodel-Large}-710M) using datasets opened by \citep{ansari2024chronos}. The training objective is auto-regressive next patch prediction with teacher forcing, based on the historical contexts.
We conduct empirical evaluations of \mymodel's zero-shot forecasting performance on 24 datasets. The experiment results demonstrate that our model achieves competitive or superior performance compared to state-of-the-art time series foundation models.

%% file: section/related.tex
\paragraph{Time Series Forecasting}
In the last decade, deep learning models have emerged as powerful tools in time series forecasting. Extensive studies have explored various architectures for building effective deep forecasting models, such as: Recurrent Neural Networks (RNNs) based models like DeepState \citep{rangapuram2018deep}, DeepAR \citep{salinas2020deepar}, ESRNN \citep{smyl2020hybrid}, and 
Convolutional Neural Networks (CNNs) based models like TCN \citep{bai2018empirical}, TimesNet \citep{wu2022timesnet}.
As Transformers \citep{vaswani2017attention} exhibited powerful sequence modeling capability and promising scalability, it has become the most popular architecture to build time series forecasting models \citep{zhou2021informer, chen2021autoformer, zhou2022fedformer, nietime, chenpathformer, ding2024drformer, liuitransformer, zhang2023crossformer}.
Some recent studies also developed linear forecasters achieving impressive performance, such as N-BEATS \citep{oreshkin2020n}, DLinear \citep{zeng2023transformers} and TiDE \citep{das2023long}. 
% From another perspective, deep forecasting models can be also categorized into two types. One is univariate models like , which predict future values solely based on past values of the single time series itself, while the other type is multivariate models leveraging cross-variable dependencies \citep{liuitransformer, zhang2023crossformer}.
While these models achieve remarkable performance, they are trained independently for each application domain and fall short in generalizability to handle cross-domain data in a wide range forecasting scenarios.

\paragraph{Time Series Foundation Models}
There have been some research efforts focusing on building time series foundation models. 
As LLMs show strong generalizability, several works adopt LLMs for time-series forecasting.
For instance, FPT \citep{zhou2023one} fine-tunes the pre-trained GPT-2 model on different time-series related tasks. 
LLMTime \citep{gruver2023large} proposes a tokenization method that encodes numerical time series as string.
Time-LLM \citep{jin2023time} aligns time series embedding to the text space via patch reprogramming and prompts LLM with aligned inputs to make future predictions.
Another line of studies concentrate on pre-training the general foundation model from scratch on a large scale of time series data. For example, ForecastFPN \citep{dooley2023forecastpfn} is a pre-trained model purely on synthetic time series and used for zero-shot forecasting.
There are also many works pre-training foundation models using real-world time series data, such as Timer \citep{liutimer}, MOIRAI \citep{woo2024unified}, Moment \citep{goswami2024moment}, and Lag-Llama \citep{rasul2023lag}, or combining real-world and synthetic data together for general pre-training like TimesFM \citep{dasdecoder}, Chronos \citep{ansari2024chronos} and TIME-MOE \citep{shi2024time}. 
For example, TimesFM \citep{dasdecoder} collects a massive amount of times series data from Google Trends and Wiki pageviews for pre-training.
% There are also many works pre-training foundation models using real-world time series data. For example, TimesFM \citep{dasdecoder} collects a massive amount of times series data from Google Trends and Wiki pageviews for pre-training.
% Other representative works includes Timer \citep{liutimer}, MOIRAI \citep{woo2024unified}, Moment \citep{goswami2024moment}, and Lag-Llama \citep{rasul2023lag}. 
% Chronos \citep{ansari2024chronos} and TIME-MOE \citep{shi2024time} combines the real-world and synthetic data together for general pre-training.
Another foundation model, TimeGPT-1 \citep{garza2023timegpt} is close-sourced and releases the commercial API for zero-shot forecasting.

%% file: section/preliminary.tex
Our goal is to build a foundation model as zero-shot time series forecaster, which can use the historical time series to predict the future value across various domains.
Formally, given the past points of a time series (also known the context) $\bm{x}_{1:C} = \{x_1, x_2, ..., x_C\}$, where $C$ is the context length, such a foundation model $f$ is expected to predict the future $H$ time points: $f:(\bm{x}_{1:C}) \longrightarrow \bm{x}_{C+1:C+H}$. 
In this work, we focus on \textit{univariate} forecasting, where $x_i$ is a scalar. 
For \textit{multivariate} time series data, the univariate model can still be applied by performing channel-independent forecasting for each individual variate of the time series.

%% file: section/method.tex
\begin{figure*}[t]
\centering
\includegraphics[width=0.99\textwidth]{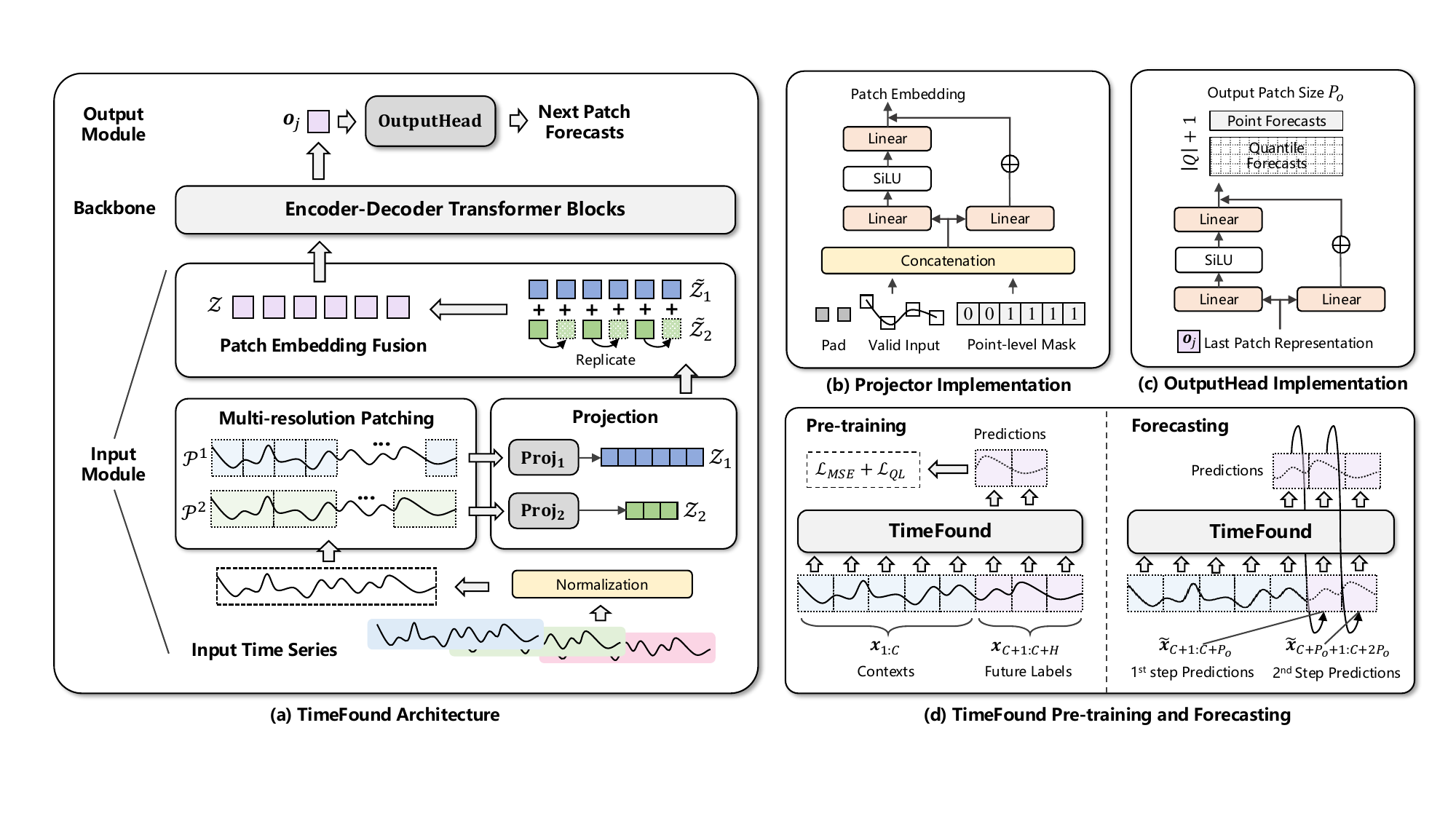}
 % \vspace{-3.5mm}
\captionsetup{font=footnotesize}
\caption{Illustration of \mymodel model. In (a), for simple illustration, we assume $K=2$ in multi-resolution patching method and it divides normalized time series using twp patch sizes $P_i$ and $P_2$. (b) and (c) show the detailed implementation of the projector in Input Module and the prediction head in Output Module respectively. (d) presents the model's different behaviors during pre-training and forecasting.}
 % \vspace{-4.5mm}
\label{fig_framework}
\end{figure*}

In this section, we will introduce \mymodel, a transformer-based foundation model for time series modeling and forecasting.
As illustrated in Figure \ref{fig_framework}, our model begins with an Input Module that pre-processes the raw time-series from different domains, where we propose a multi-resolution patching method capable in capturing the temporal dependencies at multiple scales. Next, we employ an encoder-decoder architecture, which enables both contextual understanding of historical trends and auto-regressive forecasting. 
Finally, the Output Module generates predictions of future patches. 
% Finally, the Output Module generates future predictions, supporting both point and quantile forecasting. 
Below, we provide a detailed explanation of each deigns.

\subsection{Input Module}
\paragraph{Normalization}
Since our foundation model will be pre-trained on extensive time series data with varying amplitude, frequency and stationarity, the initial step is normalizing the input data to facilitate better optimization. We apply the commonly used standard scaling method that normalize based on the mean and standard deviation calculated over the entire input series. This mitigates the bias caused by different scales across multiple datasets, while reserving the patterns of the original series.

\paragraph{Multi-resolution Patching}
The next step is to divide the time series into patches, which is analogue to the tokenization in nature language processing. Patch-based modeling has been proved to be effective in capturing the semantic information of time series data \cite{nietime} and widely adopted in recent foundation models \cite{dasdecoder}. 
Different from the vanilla patching method with fixed patch size, we propose to perform multi-resolution patch division with different patch sizes.
This design enables the model to handle time series data from different domains with distinct patterns and variations at different temporal scales, which is beneficial for building a generalizable foundation model.

Specifically, we define a collection of patch sizes $\{P_1, P_2, ..., P_K\}$, where each patch size corresponds to a division. In our framework, we restrict the value of $P_i$ to be a power of 2 and assume that $P_1 < P_2 ... < P_K$.
Given the input sequence $\bm{x}_{1:C}$, the $k$-th patch division with patch size $P_k$ will break it down into a series of $N_k$ patches $\mathcal{P}^{k} = \{\bm{p}_1^k, \bm{p}_2^k, ..., \bm{p}_{N_{k}}^k\}$, where patch $\bm{p}_{j}^{k} = \bm{x}_{(j-1)P_{k}+1:jP_{k}}$. Thus, with the multi-resolution patching method, we will obtain $K$ groups of patches: $\{\mathcal{P}^{1}, \mathcal{P}^{2}, ..., \mathcal{P}^{K}\}$.

\paragraph{Projection}
Following previous works \citep{dasdecoder, das2023long}, we encode the patches into the latent space using a two-layer Multi-layer Perceptron (MLP) projector with residual connection added to each layer. To accommodate the patches with different sizes resulted from multi-resolution patching, we employ $K$ projectors $\{\text{Proj}_{1}, \text{Proj}_{2}, ..., \text{Proj}_{K}\}$, where $\text{Proj}_{k}$ is utilized to process the $k$-th group of patches $\mathcal{P}^{k}$.

Coupled with the patch, we further introduce a point-level binary mask as a part of inputs to the projector to mark special points, such as padding. To be specific, during batch training or inference, padding is commonly adopted to fill in missing values for aligning different samples. This binary mask takes a value of 1 at valid input parts and 0 at padding positions, enabling the model to differentiate between them.
Formally, along with the input time series $\bm{x}_{1:C}$, we define the point-level mask as $\bm{m}_{1:C}$, which will be divided into patches with different sizes together. 
For the patch $\bm{p}_{j}^{k}$, the corresponding mask segment is $\bm{m}_{j}^{k}$, and they are both passed to the projector:
\begin{equation}
\bm{z}_{j}^{k} = \text{Proj}_{k}(\bm{p}_{j}^{k} \oplus \bm{m}_{j}^{k}),
\end{equation}
where $\oplus$ denotes the concatenation operation and $\bm{z}_{j}^{k}$ denotes the latent patch embedding. Thus, based on the multi-resolution patching, it yields $K$ groups of patch embeddings $\{\mathcal{Z}^{1}, \mathcal{Z}^{2}, ..., \mathcal{Z}^{K}\}$, where $\mathcal{Z}^{k} = \{\bm{z}_{1}^{k}, \bm{z}_{2}^{k}, ..., \bm{z}_{N_{k}}^{k}\}$ under the $k$-th division.

Next, we fuse the patch embeddings from different groups to form the final input for the subsequent Transformer model. 
Since different groups have varying patch sizes, they also contain different numbers of patches. To align them, we upsample the coarser groups with larger patch sizes and smaller patch numbers (i.e., $\{\mathcal{Z}_2, ..., \mathcal{Z}_K\}$ with $\{N_2, ..., N_K\}$ patches) by replicating its patches, so that they match the highest-resolution group $\mathcal{Z}_1$, which has the largest number of patches $N_1$. 
Formally, within each group $\mathcal{Z}_k$, the every patch embedding $\bm{z}_{j}^{k}$ will be repeated $N_1/N_k$ times to match the patch number of $\mathcal{Z}_1$. The replicated patch sequence can be denoted as:
\begin{equation} 
\tilde{\mathcal{Z}}^k = \{\tilde{\bm{z}}_1^k, \tilde{\bm{z}}_2^k, ..., \tilde{\bm{z}}_{N_1}^k \}, \,\, \text{where} \,\, \tilde{\bm{z}}_j^k = \bm{z}_{\lceil j\cdot N_k/N_1 \rceil}^k, \,\, j=1,2,...,N_1.
\end{equation}
After that, all groups are aligned to have the same number of patches (i.e., $N_1$), and the final patch embedding sequence $\mathcal{Z}$ is obtained by summing the corresponding patches across all groups:
\begin{equation} 
\mathcal{Z} = \{\bm{z}_1, \bm{z}_2, ..., \bm{z}_{N_1}\}, \,\, \bm{z}_j = \sum_{k=1}^{K} \tilde{\bm{z}}_j^k, 
\end{equation}
This fusion strategy ensures that information from multiple resolutions is effectively aggregated while maintaining a consistent sequence length for the subsequent Transformer processing.

\subsection{Transformer Blocks}
In our approach, we utilize the encoder-decoder architecture of T5 model \citep{raffel2020exploring} as the backbone for time series modeling and forecasting. Briefly, each block consists of a multi-head attention for contextual understanding and a feed-forward network layer for feature transformation. The relative position embedding is introduced in the calculation of attention scores.
Specifically, the encoder applies the bi-directional attention which enables to capture complex temporal dependencies among patches. While in the decoder, the causal attention is employed to ensure auto-regressive forecasting. Additionally, the decoder also integrates the cross-attention to leverage the encoded past trends and contextual information for generating predictions.
Such an encoder-decoder architecture enables the model to learn rich temporal relationships from historical data while maintaining consistency in future value generation.
Note that in the calculation of attention scores, we introduce a patch-level attention mask to filter out the padded segment, which is derived from the previously discussed point-level mask $\bm{m}_j^1$. If all values in a patch’s point-level mask are 0, it means that this patch contains only padding values, in which case the patch-level attention mask is set to 0; otherwise, it is assigned 1.

\subsection{Output Module}
Finally, an output module will project the decoder outputs into future predictions. Similar to the input module, the output module is also implemented by a two-layer MLP block with residual path. It takes the representation vector of the last patch processed by the decoder as input, and produce the prediction of the next patch. 
Formally, given the input sequence $\bm{x}_{1:C+h}$, where $\bm{x}_{1:C}$ is the context fed into the encoder, and $\bm{x}_{C+1:C+h}$ is the preceding points passed to the decoder (which can be either partial ground truth labels during training or previous predicted values during inference), both parts will be processed in a patch-wise manner.
Assuming that the last patch representation derived from the decoder is denoted as $\bm{o}_{j}$, the output module predicts the subsequent patch as follows:
\begin{equation}
    \tilde{\bm{x}}_{C+h+1: C+h+P_o} = \{\tilde{x}_{C+h+1}, \tilde{x}_{C+h+2}, ..., \tilde{x}_{C+h+P_o}\} = \text{OutputHead}(\bm{o}_{j})
\end{equation}
where $P_o$ denotes the output patch size. Note that previous studies (e.g., \citep{dasdecoder}) have indicated that a larger output patch has the advantages of improved performance and faster generation in long-term forecasting, so our approach also allows a larger output patch size than the input patch.

Though our model focuses on point forecasting, we also enable it to derive probabilistic forecasts.
% In addition to the above point forecasting, our model also supports the probabilistic forecasting. 
Following \citep{wen2017multi}, we add another prediction head to generate quantile forecasts for each time point in the next patch:
\begin{equation}
    \{\tilde{\bm{q}}_{C+h+1}, \tilde{\bm{q}}_{C+h+2}, ..., \tilde{\bm{q}}_{C+h+P_o}\} = \text{OutputHead}(\bm{o}_{j}), \,\,\,
    \text{where} \,\,
    \tilde{\bm{q}}_{i} = \{\tilde{x}_{i}^{q_1}, \tilde{x}_{i}^{q_2}, ..., \tilde{x}_{i}^{q_Q}\}
\end{equation}
where $Q$ denotes the quantile set of interest (e.g. deciles) with $q(\cdot) \in Q$, and $\tilde{x}_{i}^{q(\cdot)}$ denotes the quantile forecast value.
In the practical implementation, we use an MLP block with an output dimension of $P_o \times (|Q|+1)$ to jointly produce the point and quantiles forecasts of the next patch. And then the desired output (point or quantiles) can be obtained by slicing the results accordingly.

\subsection{Training Objective}
We pre-train \mymodel using two kinds of objectives. The first one is the commonly used Mean Squared Error (MSE) loss that minimizes the difference between point forecast values and the ground truth values. 
For an input sequence $\bm{x}_{1:C+H}$, it consists of the visible historical context $\bm{x}_{1:C}$ and the future labels $\bm{x}_{C+1:C+H}$ to be predicted.
We feed the context $\bm{x}_{1:C}$ to the encoder, pass $\bm{x}_{C+1:C+H}$ to the decoder for teacher forcing, and adopts the label shift-right method to compute the prediction error for each patch.
The loss will be computed over the entire future sequence:
\begin{equation}
    \mathcal{L}_{MSE} = \frac{1}{H} \sum_{i=C+1}^{C+H} ||x_i - \tilde{x}_i||
\end{equation}
Second, another training objective is to minimize the total Quantile Loss (QL),
based on the quantile forecasts:
% which optimizes the quantile values prediction head:
\begin{equation}
    \mathcal{L}_{QL} = \frac{1}{H} \sum_{i=C+1}^{C+H} \sum_{q \in Q} q(x_i - \tilde{x}_i^{q}) + (1-q)(\tilde{x}_i^{q} - x_i)
\end{equation}
The overall loss function is $\mathcal{L} = \mathcal{L}_{MSE} + \mathcal{L}_{QL}$, and the loss is averaged over a batch during training.

\subsection{Forecasting}
During the inference stage, our model will perform an auto-regressive forecasting in a patch-by-patch manner. Given the input time series $\bm{x}_{1:C}$ with the goal to predict the future $\bm{x}_{C+1:C+H}$, the forecasting process begins with the model generating an initial patch prediction $\tilde{\bm{x}}_{C+1:C+P_o}$. Then, this predicted $\tilde{\bm{x}}_{C+1:C+P_o}$ is fed into the decoder as part of the input to produce the next patch prediction $\tilde{\bm{x}}_{C+P_o+1:C+2P_o}$.
This process is iteratively repeated, where the model continuously integrates the predictions from the previous steps to generate subsequent patches, until the total forecasted length reaches or exceeds the target horizon length $H$.
In the case when $H$ is not an integer multiple of the output patch size $P_o$, the excess forecast points in the last patch will be discarded.

%% file: section/experiment.tex
\subsection{Pre-training Details}
\paragraph{Dataset}
To pre-train the \mymodel model, we utilized the pre-training dataset introduced by \citep{ansari2024chronos}. This dataset encompasses a diverse set of publicly available time series datasets spanning multiple domains such as energy, finance, and weather, as well as varying frequencies from five minutes to yearly. Furthermore, they applied two data augmentation strategies to enhance the diversity of the training data, where one is TSMixup strategy that generates 10M training samples by interpolating between the collected real-world time series sequences, and another is generating synthetic time series data via Gaussian processes. This ensures that the pre-training dataset can cover a broad range of forecasting scenarios. For additional details on the dataset composition and augmentation techniques, please refer to their original paper \citep{ansari2024chronos}.

\paragraph{Configuration}
We pre-train \mymodel in two sizes, namely \textit{\mymodel-Base} and  \textit{\mymodel-Large}, with key parameter details listed in Table \ref{hyper}. The models are trained for 200K steps, with a batch size of 1024. We use the AdamW optimizer with initial $lr=1e-3$, $\beta_{1}=0.9, \beta_{2}=0.999$ and $\text{weight\_decay}=0.01$. A linear learning rate decay strategy is applied over the training steps. For both models, we set the context length to 512 and the prediction length is set to 192. %We train the models on 8 NVIDIA V100-32G GPUs with TF32 precision. % 重要的参数信息，下一个版本披露

\input{table/hyper}

% 简单介绍 chronos 的数据集
% 训练超参数

% \subsection{Baselines}
% \subsection{Metrics}
% 参考 chronos

\input{table/dataset_info}

\subsection{Empirical Evaluation}
We evaluate the zero-shot forecasting performance of our model in two settings.
(1) \textbf{\textit{Standard Last Window}}. We compare all methods on the last test window of each dataset, which follows the evaluation setting for time series foundation models established in recent studies \citep{ansari2024chronos, dasdecoder, gruver2024large}.
(2) \textbf{\textit{Rolling Validation}}. This setting aims to have a more in depth understanding of our model's forecasting performance with a longer horizon. It is conducted on a subset of popular long sequence datasets, and compares the average error of the rolling validation task on the entire test set.
% 简介两种测试方法、last and cross validation

\subsubsection{Zero-shot Evaluation: Last Window Setting}
\paragraph{Setups}
We evaluate the zero-shot forecasting ability of \mymodel on 24 datasets that were unseen during the pre-training stage. Table \ref{dataset} lists the details of these datasets. 
These datasets are largely consistent with the benchmark II introduced in the \citep{ansari2024chronos}, which comprises 27 datasets, mostly from the Monash \citep{godahewa2021monash} and Informer \citep{zhou2021informer}, except for the $ERCOT$ $Load$ and $Exchange$ $Rate$ datasets. The key difference is that we have removed three datasets that were included in the pre-training data of baseline models. These datasets covers multiple domains and granularities.
For each dataset, we report the errors on the last test window. The forecast horizon is determined according to the sampling frequency.

We compare the performance of our model against two recent state-of-the-art foundation models for zero-shot time series forecasting, Chornos \citep{ansari2024chronos} and TimesFM \citep{dasdecoder}.
The evaluation metrics include the Mean Absolute Scaled Error (MASE, \cite{hyndman2006another}) and symmetric Mean Absolute Percentage Error (sMAPE).
Since the magnitude of metrics vary across datasets, we follow \cite{ansari2024chronos} to compute the relative score of each model relative to a baseline approach, Seasonal Naive, on each dataset. Then, we report the geometric mean of the relative scores across all datasets as the overall performance of the model.

\input{table/exp_benchmark}

% 数据集 参数
\paragraph{Results}
The results are shown in Table \ref{table-exp-benchmark}. As we can see, the proposed \mymodel model demonstrates strong zero-shot forecasting performance.
Notably, our model achieves the best average performance (geometric mean of MASE), compared with the state-of-the-art foundation models TimesFM and Chornos, which indicates its good generalization ability across diverse forecasting scenarios. 
Also, our model can rank the first or the second on most datasets, particularly in the MASE metric. 
While TimesFM attains the highest ranking on a slightly larger number of individual datasets, its accuracy on other datasets falls significantly behind other approaches, leading to a lower average performance. 
Given that this benchmark only evaluated the model on the last test window, the statistical significance of individual dataset results is limited, thus the geometric mean result is a more reliable measure for evaluating overall performance. Moreover, as a foundation model, it is crucial to achieve strong results across diverse forecasting scenarios rather than excelling on just a few datasets. In this regard, geometric mean serves as a better indicator of a model’s generalization ability. Therefore, \mymodel with highest geometric mean results is considered to have superior overall effectiveness.

\subsubsection{Long-horizon Zero-shot Evaluation: Rolling Validation Setting}
\paragraph{Setups}
We further benchmark our models' long-horizon forecasting ability on four electricity transformer temperature datasets (ETTh$_1$, ETTh$_2$, ETTm$_1$, ETTm$_2$) collected by \cite{zhou2021informer}. In this setting, each model performs rolling forecasting across the entire test set.
We compare their performance on horizon lengths of \{96, 192, 336 and 720\}, while the context length is fixed at 512.

In this experiment, another state-of-the-art foundation model Timer \citep{liutimer} is included for comparison. This model is not compared in the \textit{Last Window} setting because most of the test datasets have been seen during its pre-training stage.
% We select the Mean Squared Error (MSE) and Mean Absolute Error (MAE) as evaluation metrics, 
We select the Mean Absolute Error (MAE) and symmetric Mean Absolute Percentage Error (sMAPE) as metrics, 
and report the results calculated on the standard normalized data.

% Therefore following [GFQW23] we scale the metric of each baseline for a dataset by the same metric achieved by a naive baseline on that dataset. The naive baseline just makes the constant prediction yL repeated across the prediction length. We did not need to do that for the Informer datasets since on these datasets metrics are usually reported on standard normalized data [NNSK22].

% \input{table/exp_rolling}
\input{table/exp_rolling1}

\paragraph{Results}
Table \ref{table-exp-rolling} presents the results, where we report the results of \textit{Large} model for Chronos and our \mymodel. It shows that our model achieves the best overall performance in the long-horizon zero-shot forecasting task on these datasets, and it can consistently deliver strong results across different horizon lengths. 
We also observe that the Chronos baseline shows relatively poor performance. This is likely because it adopts point-based modeling for time series data (while other approaches are patch-based), and auto-regressively generates future predictions in a point-by-point manner, which leads to significant error accumulation in long-horizon forecasting. 
This provides valuable insights that patch-based modeling and prediction are crucial for building strong time series foundation models, particularly when the model is applied in long-horizon forecasting tasks.

%% file: table/hyper.tex
\renewcommand\arraystretch{1.0}
\begin{table}[ht]
\centering
\caption{Configuration of \mymodel model.}
\resizebox{1.0\textwidth}{!}{
    \begin{tabular}{cccccccc}
    \toprule
    & Model Size & \# Encoder Layers & \# Decoder Layers & Hidden Size & \# Heads & Patch Size & Output Patch Size \\
    Base & 200M & 12 & 12 & 768 & 12 & $\{P_1=16, P_2=32\}$ & $P_o=32$ \\
    Large & 710M & 24 & 24 & 1024 & 16 & $\{P_1=16, P_2=32\}$ & $P_o=32$ \\
    \bottomrule
    \end{tabular}
    }
\label{hyper}
\end{table}

%% file: table/dataset_info.tex
\renewcommand\arraystretch{1.0}
\begin{table}[ht]
\centering
\caption{Details of zero-shot evaluation datasets. This table is modified from \cite{ansari2024chronos}.}
\resizebox{0.95\textwidth}{!}{
\begin{tabular}{lccrrrr}
\toprule
\textbf{Dataset} & \textbf{Domain} & \textbf{Frequency} & \textbf{Num. Series} & \textbf{Min. Length} & \textbf{Max. Length} & \textbf{Horizon Length} \\
\midrule
Australian Electricity & Energy & 30min & 5 & 230736  & 232272  & 60 \\
CIF 2016 & Banking & 1M & 72 & 28  & 120  & 12 \\
Car Parts & Retail & 1M & 2674 & 51  & 51  & 12 \\
Covid Deaths & Healthcare & 1D & 266 & 212  & 212  & 30 \\
Dominick & Retail & 1D & 100014 & 201  & 399  & 8 \\
ERCOT Load & Energy & 1H & 8 & 154854  & 154854  & 24 \\
ETT (15 Min.) & Energy & 15min & 14 & 69680  & 69680  & 24 \\
ETT (Hourly) & Energy & 1H & 14 & 17420  & 17420  & 24 \\
Exchange Rate & Finance & 1B & 8 & 7588  & 7588  & 30 \\
FRED-MD & Economics & 1M & 107 & 728  & 728  & 12 \\
Hospital & Healthcare & 1M & 767 & 84  & 84  & 12 \\
M1 (Monthly) & Various & 1M & 617 & 48  & 150  & 18 \\
M1 (Quarterly) & Various & 3M & 203 & 18  & 114  & 8 \\
M1 (Yearly) & Various & 1Y & 181 & 15  & 58  & 6 \\
M3 (Monthly) & Various & 1M & 1428 & 66  & 144  & 18 \\
M3 (Quarterly) & Various & 3M & 756 & 24  & 72  & 8 \\
M3 (Yearly) & Various & 1Y & 645 & 20  & 47  & 6 \\
M5 & Retail & 1D & 30490 & 124  & 1969  & 28 \\
NN5 (Daily) & Finance & 1D & 111 & 791  & 791  & 56 \\
NN5 (Weekly) & Finance & 1W & 111 & 113  & 113  & 8 \\
Tourism (Monthly) & Various & 1M & 366 & 91  & 333  & 24 \\
Tourism (Quarterly) & Various & 1Q & 427 & 30  & 130  & 8 \\
Tourism (Yearly) & Various & 1Y & 518 & 11  & 47  & 4 \\
Weather & Nature & 1D & 3010 & 1332  & 65981  & 30 \\

    \bottomrule
    \end{tabular}
    }
\label{dataset}
\end{table}

%% file: table/exp_benchmark.tex
\renewcommand\arraystretch{1.0}
\begin{table}[t]
\centering
\captionsetup{font=footnotesize}
\caption{Comparison of zero-shot forecasting performance under the \textit{standard last window} setting. The best (second best) results are in red (blue). $*$Note$*$: for Chronos variants, we report the results on the median trajectory of 20 sampled trajectories.}
\vspace{1mm}
\resizebox{1.0\textwidth}{!}{
    \begin{tabular}{l|c|c|c|c|c|c|c|c|c|c}
    \toprule
    & \multicolumn{5}{c|}{MASE $\downarrow$} & \multicolumn{5}{c}{sMAPE $\downarrow$} \\
    % \cline{2-11}
    % \rule{0pt}{11pt}
    \cmidrule{2-11}
    % & TimesFM & Chornos-Base & Chornos-Large & \mymodel-Base & \mymodel-Large & TimesFM & Chornos-Base & Chornos-Large & \mymodel-Base & \mymodel-Large \\
    & TimesFM & Chornos & Chornos & \mymodel & \mymodel & TimesFM & Chornos & Chornos & \mymodel & \mymodel \\
    & & (Base) & (Large) & (Base) & (Large) & & (Base) & (Large) & (Base) & (Large) \\
    \midrule
    Australian Electricity & \textcolor{blue}{1.089} & 1.141 & 1.273 & \textbf{\textcolor{red}{1.044}} & 1.177 & \textcolor{blue}{0.048} & 0.049 & 0.053 & \textbf{\textcolor{red}{0.047}} & 0.049 \\
    Car Parts & 0.841 & 0.811 & \textbf{\textcolor{red}{0.806}} & 0.821 & \textcolor{blue}{0.807} & \textbf{\textcolor{red}{0.934}} & 0.951 & 0.948 & 0.942 & \textcolor{blue}{0.941} \\
    CIF 2016 & 1.036 & 0.992 & \textcolor{blue}{0.979} & 1.013 & \textbf{\textcolor{red}{0.971}} & 0.071 & 0.073 & \textcolor{blue}{0.071} & \textbf{\textcolor{red}{0.070}} & 0.074 \\
    Covid Deaths & 7.803 & 6.409 & 6.518 & \textbf{\textcolor{red}{5.486}} & \textcolor{blue}{6.020} & 0.228 & 0.201 & 0.205 & \textbf{\textcolor{red}{0.190}} & \textcolor{blue}{0.197} \\
    Dominick & 0.938 & \textbf{\textcolor{red}{0.770}} & \textcolor{blue}{0.774} & 0.868 & 0.850 & 0.791 & 0.810 & 0.810 & \textbf{\textcolor{red}{0.789}} & \textcolor{blue}{0.790} \\
    ERCOT Load & \textcolor{blue}{0.589} & \textbf{\textcolor{red}{0.567}} & 0.635 & 0.616 & 0.636 & \textcolor{blue}{0.011} & \textbf{\textcolor{red}{0.011}} & 0.012 & 0.011 & 0.012 \\
    ETT (15 Min.) & \textbf{\textcolor{red}{0.602}} & \textcolor{blue}{0.648} & 0.761 & 0.682 & 0.679 & \textbf{\textcolor{red}{0.093}} & \textcolor{blue}{0.101} & 0.116 & 0.103 & 0.104 \\
    ETT (Hourly) & 0.890 & 0.779 & \textbf{\textcolor{red}{0.758}} & 0.789 & \textcolor{blue}{0.777} & 0.098 & \textbf{\textcolor{red}{0.090}} & \textcolor{blue}{0.091} & 0.096 & 0.097 \\
    Exchange Rate & 1.698 & 2.103 & 1.954 & \textcolor{blue}{1.605} & \textbf{\textcolor{red}{1.494}} & \textcolor{blue}{0.005} & 0.006 & 0.006 & 0.005 & \textbf{\textcolor{red}{0.005}} \\
    FRED-MD & 0.650 & \textbf{\textcolor{red}{0.495}} & \textcolor{blue}{0.520} & 0.565 & 0.573 & 0.056 & \textbf{\textcolor{red}{0.049}} & \textcolor{blue}{0.050} & 0.052 & 0.052 \\
    Hospital & \textbf{\textcolor{red}{0.783}} & 0.815 & 0.809 & 0.798 & \textcolor{blue}{0.792} & \textbf{\textcolor{red}{0.089}} & 0.094 & 0.093 & \textcolor{blue}{0.090} & 0.090 \\
    M1 (Monthly) & \textbf{\textcolor{red}{1.068}} & 1.131 & \textcolor{blue}{1.100} & 1.139 & 1.103 & \textbf{\textcolor{red}{0.075}} & 0.079 & \textcolor{blue}{0.077} & 0.079 & 0.077 \\
    M1 (Quarterly) & \textbf{\textcolor{red}{1.671}} & 1.768 & \textcolor{blue}{1.706} & 1.749 & 1.714 & \textbf{\textcolor{red}{0.079}} & 0.090 & 0.086 & 0.085 & \textcolor{blue}{0.085} \\
    M1 (Yearly) & \textbf{\textcolor{red}{4.004}} & 4.462 & 4.413 & \textcolor{blue}{4.203} & 4.304 & \textbf{\textcolor{red}{0.097}} & 0.112 & 0.110 & \textcolor{blue}{0.104} & 0.106 \\
    M3 (Monthly) & 0.935 & 0.872 & \textcolor{blue}{0.866} & 0.890 & \textbf{\textcolor{red}{0.864}} & 0.073 & \textcolor{blue}{0.070} & \textbf{\textcolor{red}{0.070}} & 0.072 & 0.071 \\
    M3 (Quarterly) & \textbf{\textcolor{red}{1.151}} & 1.214 & 1.198 & 1.232 & \textcolor{blue}{1.190} & \textcolor{blue}{0.049} & 0.051 & \textbf{\textcolor{red}{0.049}} & 0.051 & 0.050 \\
    M3 (Yearly) & \textbf{\textcolor{red}{2.697}} & 3.178 & 3.070 & 2.999 & \textcolor{blue}{2.931} & \textbf{\textcolor{red}{0.080}} & 0.092 & 0.089 & 0.089 & \textcolor{blue}{0.087} \\
    M5 & \textbf{\textcolor{red}{1.397}} & 1.430 & 1.430 & 1.412 & \textcolor{blue}{1.412} & \textbf{\textcolor{red}{0.778}} & 0.820 & 0.820 & \textcolor{blue}{0.788} & 0.794 \\
    NN5 (Daily) & 0.894 & \textcolor{blue}{0.843} & \textbf{\textcolor{red}{0.833}} & 0.875 & 0.866 & 0.112 & \textcolor{blue}{0.105} & \textbf{\textcolor{red}{0.104}} & 0.110 & 0.108 \\
    NN5 (Weekly) & 0.949 & \textbf{\textcolor{red}{0.932}} & 0.949 & \textcolor{blue}{0.935} & 0.937 & 0.058 & \textbf{\textcolor{red}{0.058}} & 0.059 & \textcolor{blue}{0.058} & 0.058 \\
    Tourism (Monthly) & 1.918 & 1.861 & 1.819 & \textcolor{blue}{1.663} & \textbf{\textcolor{red}{1.610}} & 0.122 & 0.128 & 0.125 & \textcolor{blue}{0.107} & \textbf{\textcolor{red}{0.104}} \\
    Tourism (Quarterly) & 2.063 & 1.782 & \textbf{\textcolor{red}{1.649}} & \textcolor{blue}{1.669} & 1.746 & 0.099 & 0.088 & \textbf{\textcolor{red}{0.082}} & \textcolor{blue}{0.082} & 0.087 \\
    Tourism (Yearly) & \textbf{\textcolor{red}{3.233}} & 3.895 & \textcolor{blue}{3.686} & 3.961 & 3.808 & \textbf{\textcolor{red}{0.181}} & 0.232 & \textcolor{blue}{0.213} & 0.239 & 0.235 \\
    Weather & 0.627 & 0.561 & 0.565 & \textcolor{blue}{0.559} & \textbf{\textcolor{red}{0.546}} & 0.320 & 0.331 & 0.330 & \textbf{\textcolor{red}{0.305}} & \textcolor{blue}{0.311} \\
    \midrule
    Geometric Mean & 0.869 & 0.857 & 0.859 & \textcolor{blue}{0.846} & \textbf{\textcolor{red}{0.842}} & \textbf{\textcolor{red}{1.105}} & 1.133 & 1.130 & \textcolor{blue}{1.109} & 1.109 \\
    \bottomrule
    \end{tabular}
    }
\label{table-exp-benchmark}
\end{table}

%% file: table/exp_rolling1.tex
\renewcommand\arraystretch{1.0}
\begin{table*}[t]
    \captionsetup{font=footnotesize}
    \caption{Comparison of \textit{long-horizon} zero-shot forecasting performance under the \textit{rolling validation} setting. The best (second best) results are in red (blue). $*$Note$*$: for Chronos variants, we report the results on the median trajectory of 20 sampled trajectories.}
    % \vspace{1mm}
    % \small
    % \footnotesize
    \centering
    \resizebox{0.9\textwidth}{!}{
	\begin{tabular}{c|c|c|c|c|c|c|c|c|c}
	\toprule
         % & & \multicolumn{4}{c|}{MSE $\downarrow$} & \multicolumn{4}{c}{MAE $\downarrow$} \\
        & & \multicolumn{4}{c|}{MAE $\downarrow$} & \multicolumn{4}{c}{sMAPE $\downarrow$} \\
        % \cline{1-10}
        % \rule{0pt}{12pt}
        \cmidrule{1-10}
        Dataset & Horizon & Timer & TimesFM & Chornos & \mymodel & Timer & TimesFM  & Chornos & \mymodel \\
        % & & & & (Large) & (Large) & &  & (Large) & (Large) \\
        % Dataset & Horizon & Timer & TimesFM & Chornos (Large) & \mymodel (Large) & Timer & TimesFM  & Chornos (Large) & \mymodel (Large) \\
        \midrule
        
\multirow{5}{*}{ETTh1}
& 96 & \textcolor{blue}{0.388} & 0.405 & 0.404 & \textbf{\textcolor{red}{0.384}} & 0.721 & 0.725 & \textcolor{blue}{0.719} & \textbf{\textcolor{red}{0.707}} \\
& 192 & \textbf{\textcolor{red}{0.411}} & 0.432 & 0.451 & \textcolor{blue}{0.417} & \textcolor{blue}{0.743} & 0.758 & 0.772 & \textbf{\textcolor{red}{0.737}} \\
& 336 & \textbf{\textcolor{red}{0.431}} & 0.459 & 0.468 & \textcolor{blue}{0.442} & \textcolor{blue}{0.753} & 0.795 & 0.805 & \textbf{\textcolor{red}{0.741}} \\
& 720 & \textcolor{blue}{0.471} & 0.482 & 0.521 & \textbf{\textcolor{red}{0.452}} & \textcolor{blue}{0.856} & 0.888 & 0.906 & \textbf{\textcolor{red}{0.812}} \\
& average & \textcolor{blue}{0.425} & 0.444 & 0.461 & \textbf{\textcolor{red}{0.423}} & \textcolor{blue}{0.768} & 0.792 & 0.800 & \textbf{\textcolor{red}{0.749}} \\
\midrule
\multirow{5}{*}{ETTh2}
& 96 & 0.342 & 0.344 & \textcolor{blue}{0.337} & \textbf{\textcolor{red}{0.329}} & 0.549 & 0.516 & \textbf{\textcolor{red}{0.509}} & \textcolor{blue}{0.511} \\
& 192 & 0.399 & \textcolor{blue}{0.393} & \textbf{\textcolor{red}{0.383}} & 0.394 & 0.617 & \textbf{\textcolor{red}{0.581}} & 0.587 & \textcolor{blue}{0.586} \\
& 336 & 0.405 & 0.403 & \textbf{\textcolor{red}{0.398}} & \textcolor{blue}{0.400} & 0.630 & \textcolor{blue}{0.617} & 0.631 & \textbf{\textcolor{red}{0.604}} \\
& 720 & \textbf{\textcolor{red}{0.430}} & 0.484 & 0.485 & \textcolor{blue}{0.474} & 0.693 & 0.712 & \textbf{\textcolor{red}{0.637}} & \textcolor{blue}{0.666} \\
& average & \textbf{\textcolor{red}{0.394}} & 0.406 & 0.401 & \textcolor{blue}{0.399} & 0.622 & 0.606 & \textbf{\textcolor{red}{0.591}} & \textcolor{blue}{0.592} \\
\midrule
\multirow{5}{*}{ETTm1}
& 96 & 0.369 & \textcolor{blue}{0.351} & 0.378 & \textbf{\textcolor{red}{0.320}} & 0.689 & \textcolor{blue}{0.675} & 0.708 & \textbf{\textcolor{red}{0.619}} \\
& 192 & 0.400 & \textcolor{blue}{0.390} & 0.434 & \textbf{\textcolor{red}{0.358}} & 0.729 & \textcolor{blue}{0.726} & 0.794 & \textbf{\textcolor{red}{0.659}} \\
& 336 & 0.426 & \textcolor{blue}{0.420} & 0.469 & \textbf{\textcolor{red}{0.384}} & 0.780 & \textcolor{blue}{0.778} & 0.875 & \textbf{\textcolor{red}{0.702}} \\
& 720 & 0.490 & \textcolor{blue}{0.472} & 0.524 & \textbf{\textcolor{red}{0.443}} & 0.883 & \textcolor{blue}{0.850} & 0.956 & \textbf{\textcolor{red}{0.784}} \\
& average & 0.421 & \textcolor{blue}{0.408} & 0.451 & \textbf{\textcolor{red}{0.376}} & 0.770 & \textcolor{blue}{0.757} & 0.833 & \textbf{\textcolor{red}{0.691}} \\
\midrule
\multirow{5}{*}{ETTm2}
& 96 & 0.274 & \textcolor{blue}{0.257} & 0.262 & \textbf{\textcolor{red}{0.246}} & 0.466 & \textcolor{blue}{0.435} & 0.447 & \textbf{\textcolor{red}{0.415}} \\
& 192 & 0.313 & 0.314 & \textcolor{blue}{0.296} & \textbf{\textcolor{red}{0.288}} & 0.505 & 0.486 & \textcolor{blue}{0.477} & \textbf{\textcolor{red}{0.450}} \\
& 336 & \textcolor{blue}{0.365} & 0.397 & 0.378 & \textbf{\textcolor{red}{0.360}} & 0.549 & \textcolor{blue}{0.544} & 0.549 & \textbf{\textcolor{red}{0.511}} \\
& 720 & \textcolor{blue}{0.412} & 0.445 & 0.446 & \textbf{\textcolor{red}{0.408}} & \textcolor{blue}{0.596} & 0.604 & 0.608 & \textbf{\textcolor{red}{0.565}} \\
& average & \textcolor{blue}{0.341} & 0.353 & 0.345 & \textbf{\textcolor{red}{0.326}} & 0.529 & \textcolor{blue}{0.517} & 0.520 & \textbf{\textcolor{red}{0.486}} \\

\bottomrule
\end{tabular}
}
% \vspace{-2mm}
\label{table-exp-rolling}
\end{table*}

%% file: section/conclusion.tex
In this paper, we introduced \mymodel, a foundation model designed for zero-shot time series forecasting across various domains.
Our model employs a multi-resolution patching strategy within an encoder-decoder transformer architecture to capture complex temporal dynamics across different scales from a diverse set of time series data.
Our experiments show that \mymodel achieves promising zero-shot results, superior to or competitive with the state-of-the-art time series foundation models, on multiple unseen datasets.